\documentclass[conference,a4paper]{IEEEtran}
\usepackage[T5]{fontenc}
\usepackage[utf8]{inputenc}
\usepackage[vietnamese,english]{babel}

\usepackage{graphicx}
\usepackage{amsmath}
\usepackage{amssymb}
\usepackage{cite}

\usepackage{booktabs}
\usepackage{tabularx}
\usepackage{array}
\usepackage{multirow}
\usepackage{float}
\usepackage{hyperref}
\hypersetup{
    colorlinks=true,
    linkcolor=blue,
    citecolor=blue,
    urlcolor=blue
}

\usepackage{xcolor}
\usepackage[table]{xcolor}
\definecolor{myblue}{RGB}{0, 51, 102}

\usepackage{algorithm}
\usepackage{algpseudocode}

\begin{document}

\title{DynaCalKV: Key-Value Cache Compression via Head Grouping and Adaptive Rank Allocation}

\author{
  \IEEEauthorblockN{
    Tan T. Nguyen\IEEEauthorrefmark{1}, and 
    Quan V. Dang\IEEEauthorrefmark{1},\IEEEauthorrefmark{2}
  }
  \IEEEauthorblockA{
    \IEEEauthorrefmark{1}Full Stack Data Science\\
    \IEEEauthorrefmark{2}Department of Computer Science, University College London\\
    Emails: nguyenthaitan02@gmail.com,  dangvanquan.nd@gmail.com
  }
}

\maketitle

\begin{abstract}
As the inference phase of Large Language Models (LLMs) requires handling long context windows, the Key-Value (KV) cache initially appears to address this challenge but eventually becomes a significant bottleneck as the context window continues to grow. Low-rank compression has recently been studied as an effective approach to reduce KV cache memory while maintaining model performance. However, only a few existing methods treat the Key and Value caches differently, despite their distinct roles. Moreover, these methods typically employ fixed attention-head grouping, which may not fully exploit the structural similarity among attention heads. In this paper, we propose an improved low-rank KV cache compression framework. For the Key cache, we dynamically group attention heads based on Centered Kernel Alignment (CKA) similarity and allocate the rank budget adaptively under a parameter budget. For the Value cache, we adopt the same approach as ReCalKV, refining the low-rank decomposition through offline calibration to improve reconstruction quality. Experimental results on three instruction-tuned LLMs show that our method reduces the number of Key cache parameters while maintaining competitive accuracy. We further observe that the proposed strategy is particularly effective for Multi-Head Attention (MHA) models, whereas it should be applied more conservatively to Grouped-Query Attention (GQA) models, especially in long-context settings.
\end{abstract}


\section{Introduction} 

Large Language Models (LLMs) have recently demonstrated a significant role across almost all natural language tasks. A considerable factor in the success of these LLMs is the context window, which encompasses all the information, including prompts, chat histories, and uploaded documents, that the LLMs can read, remember, and process to generate a response. As real-world language tasks become more complex, the context window must also become larger, requiring the capability to handle long-context windows. As a result, memory consumption during inference has become a critical bottleneck. One of the most famous approaches to addressing this is the Key-Value (KV) cache, which stores the attention representations of previously generated tokens. By doing so, it avoids repeated computation; however, the memory still grows linearly with the sequence length, becoming a new obstacle.

To handle this issue, numerous KV cache compression methods have been proposed. Previous approaches can generally be categorized into quantization-based methods, token eviction methods, and low-rank compression methods. Among these, low-rank approaches are particularly attractive because they preserve the entire context while substantially reducing memory usage. Nevertheless, only a few recent works focus on the differences between the behaviors of Key and Value caches, which require distinct compression strategies; moreover, these existing methods still rely on simplifying assumptions. For instance, ReCalKV \cite{yan2025recalkv} groups attention heads using a fixed grouping strategy and allocates the rank budget uniformly across all groups. However, attention heads may exhibit varying degrees of similarity, making fixed grouping suboptimal and motivating a dynamic grouping strategy with adaptive rank allocation.

In this paper, we propose an improved low-rank KV cache compression framework that answers these questions. For the Key cache, we dynamically group attention heads based on their similarity measured by Centered Kernel Alignment (CKA), and subsequently allocate the rank budget adaptively according to the importance of each group, while guaranteeing that the number of parameters does not exceed that of previous methods. For the Value cache, we decompose the entire original projection matrix rather than processing it in groups, and then refine the decomposed factors through another procedure that minimizes the reconstruction error on calibration data. Various experiments on language models indicate that the effectiveness of dynamic head grouping depends on the underlying attention architecture. While the proposed method is highly suitable for standard MHA models, it should be applied more conservatively to GQA models in long-context settings.

The main contributions of this work are summarized as follows:

\begin{itemize}
    \item We propose DynaCalKV, a dynamic Key cache compression framework based on CKA similarity and a subsequent adaptive rank allocation algorithm, which preserves the parameter budget comparing to the previous methods.
    
    \item Experimental results on multiple LLMs and datasets show that DynaCalKV improves memory efficiency while maintaining competitive performance, and reveal that the proposed strategy is particularly effective for MHA models but should be applied more cautiously to GQA models.
\end{itemize}

\section{Related Work}
\textbf{KV Cache Compression.} To reduce the memory required for the KV cache during LLM inference, especially in long-context scenarios, numerous methods have been proposed. Quantization-based methods, such as KIVI \cite{liu2024kivi}, KVQuant \cite{hooper2024kvquant}, aim to reduce the size of KV cache representations by using lower-bit numerical formats. Token eviction methods, such as H$_2$O \cite{zhang2023h2o}, SnapKV \cite{li2024snapkv}, CAKE \cite{qin2025cake}, discard tokens that are deemed unimportant or redundant. Notably, these two methods can be applied orthogonally to another group of methods, low-rank compression methods. These methods reduce the dimensionality of KV representations through low-rank projections or matrix decomposition techniques.

\textbf{Low-Rank KV Cache Compression.} Among those categories, quantization-based and token eviction approaches have received more attention, whereas low-rank compression methods represent a promising research direction with considerable potential for future development. MatryoshkaKV \cite{lin2025matryoshkakv} and Eigen-Attention \cite{saxena2024eigen} learn additional projection matrices to map the KV cache into a lower-dimensional latent space. In contrast, LoRC \cite{zhang2024lorc} and Palu \cite{chang2024palu} directly utilize singular value decomposition (SVD) to represent the KV cache using smaller matrices instead of the original one. These studies reveal the effectiveness of low-rank compression for KV cache reduction.

\textbf{Different strategies for Key and Value in Low-Rank KV Cache Compression.} As research in this area continues to evolve, a promising approach that has not yet been broadly explored is to treat the Key cache and Value cache differently, given their distinct roles within the transformer architecture. Most studies following this line of research share the observation that the Key cache can be compressed more aggressively than the Value cache, based on the assumption that the Key primarily serves as an index for attention, whereas the Value contains the complete information. Both AsymKV \cite{cui2026homogeneous} and ReCalKV \cite{yan2025recalkv} propose compressing the Key cache using static groups while handling the Value cache without grouping. In a different approach, Thin Keys, Full Values \cite{yao2026thin} focuses only on the Key cache while leaving the Value cache completely untouched. Unlike prior works, our method introduces dynamic grouping for the Key cache while also compressing the Value cache instead of leaving it unmodified.















\section{Methodology}

\subsection{Preliminary}

\textbf{Singular Value Decomposition.} Singular Value Decomposition (SVD) is a popular matrix factorization method, which is widely adopted in constructing low-rank approximations of matrices. For any matrix $W \in \mathbb{R}^{m \times n}$, SVD decomposes it into three matrices, $W = U \Sigma V^\top$, where $U \in \mathbb{R}^{m \times m}$ and $V \in \mathbb{R}^{n \times n}$ are orthogonal matrices containing the left and right singular vectors, respectively, while $\Sigma \in \mathbb{R}^{m \times n}$ is a diagonal matrix whose entries are the singular values. Given $r \ll \min(m,n)$, to obtain a rank-$r$ approximation, only the top-$r$ singular values and vectors are retained, yields $W \approx LR$, where $L = U_r \Sigma_r^{1/2}$, $R = \Sigma_r^{1/2} V_r^\top$, with $U_r \in \mathbb{R}^{m \times r}$, $V_r \in \mathbb{R}^{n \times r}$, and $\Sigma_r \in \mathbb{R}^{r \times r}$ correspond to the leading $r$ singular components. For an input $x \in \mathbb{R}^{1 \times m}$, we cache the compressed representation $z=xL \in \mathbb{R}^{1 \times r}$ instead of $xW \in \mathbb{R}^{1 \times n}$, and reconstruct the projection as $xW \approx zR$, reducing KV cache memory with a compression ratio of $r/n$.

\textbf{Centered Kernel Alignment Similarity.} Centered Kernel Alignment (CKA) is a similarity metric for comparing the representations learned by different neural network layers or models. Given two centered representation matrices $X \in \mathbb{R}^{n \times d_x}$ and $Y \in \mathbb{R}^{n \times d_y}$, the linear CKA similarity is defined as
\[\mathrm{CKA}(X,Y) = \dfrac{\|Y^\top X\|_F^2}{\|X^\top X\|_F\|Y^\top Y\|_F},\]
where $\|\cdot\|_F$ denotes the Frobenius norm. The CKA score ranges from $0$ to $1$, where a larger value indicates greater similarity between the two representations.


\subsection{Compression ratio allocation}
Driven by Palu \cite{chang2024palu}, we evaluate the relative importance of each layer by computing layer-wise Fisher Information scores on the calibration data. These scores are subsequently utilized to allocate the rank budget across layers.

For a given layer and its allocated rank budget, we apply distinct compression strategies depending on whether it is a Key or Value layer, as detailed in the following subsections.

\subsection{Key compression}
Given a Key projection matrix $W^{(k)} \in \mathbb{R}^{m \times n}$, where $n=h\cdot d_h$ corresponds to $h$ attention heads each with a hidden dimension of $d_h$, and a total rank budget $r$ for this matrix, we aim to further allocate the rank to groups of heads. Suppose $W^{(k)} \in \mathbb{R}^{m \times n}$ is partitioned column-wise into $K$ submatrices, where each submatrix represents a distinct group of heads. Formally, those submatrices are denoted as $W_i \in \mathbb{R}^{m \times h_i d_h}$ for $i \in \{1,\dots,K\}$, where $h_i$ represents the number of heads assigned to the $i$-th group such that $\sum_{i=1}^K h_i = h$. The head ratio for group $i$ is defined as $\alpha_i = h_i/h$. For each group $i$, a low-rank approximation $W_i \approx L_i R_i$ is performed, where $L_i \in \mathbb{R}^{m \times r_i}$ and $R_i \in \mathbb{R}^{r_i \times h_i d_h}$. Here, $r_i$ denotes the rank allocated to the $i$-th group, satisfying the constraint $\sum_{i=1}^K r_i = r$. The total number of parameters used for each $W^{(k)}$ after decomposition is then 
\[\begin{aligned}
\sum_{i=1}^K \left[ (m \cdot r_i) + (r_i \cdot h_i d_h) \right] = 
mr + n \left(\sum_{i=1}^K r_i\alpha_i\right).
\end{aligned}\]

The next step is to determine how to partition $W^{(k)}$. Following ReCalKV \cite{yan2025recalkv}, we first measure the pairwise similarity between attention heads to identify those with closely related structures. Specifically, each entry $S_{i,j}$ of the similarity matrix $S \in \mathbb{R}^{h \times h}$, where $h$ denotes the number of attention heads, represents the similarity between $i$-th head and $j$-th head. In our implementation, we utilize Centered Kernel Alignment (CKA), which measures representational similarity by comparing the feature representations produced by different heads, making it suitable for identifying heads that encode similar information.

After obtaining the similarity matrix $S$, ReCalKV \cite{yan2025recalkv} reorders the attention heads and partitions them into groups of a fixed size, specifically four heads per group. Moreover, all groups are compressed using the same compression ratio. In other words, in this case, $\alpha_i = 1/K$, $r_i = r/K$ for all $i$, and $K = h/4$. Under this uniform allocation configuration, the total number of parameters required to represent $W^{(k)}$ after low-rank decomposition is
\[\begin{aligned}
mr + rn \dfrac{4}{h}.
\end{aligned}\]

\begin{algorithm}[t]
\caption{Key Compression for a layer}
\label{algo:key_compression}
\begin{algorithmic}[1]

\State \textbf{Input:} Projection matrix $W^{(k)}$, similarity matrix $S$, rank budget $r$, candidates $\mathcal{K}$
\State \textbf{Output:} Optimal factors $\{L_i, R_i\}$

\Procedure{Compress}{$W^{(k)}, S, r, \mathcal{K}$}
    \ForAll{$K \in \mathcal{K}$}
        \State $\{W_i\} \gets \Call{ClusterHeads}{S, K}$
        \State $\{r_i\} \gets \Call{InitRanksByEnergy}{\{W_i\}, r}$
        \State $\{r_i\} \gets \Call{GreedyAdjust}{\{r_i\}, r}$
        \State $\{L_i, R_i\} \gets \Call{GroupDecompose}{\{W_i\}, \{r_i\}}$
        \State $\mathcal{L}(K) \gets \Call{ComputeError}{\{W_i\}, \{L_i, R_i\}, \{r_i\}}$
    \EndFor
    \State $K^* \gets \arg\min_{K \in \mathcal{K}} \mathcal{L}(K)$
    \State \Return $\{L_i, R_i\}$ corresponding to $K^*$
\EndProcedure

\end{algorithmic}
\end{algorithm}

Instead of enforcing a predetermined number of heads per group as ReCalKV, in DynaCalKV, groups are now formed adaptively according to $S$. This dynamic grouping strategy better captures the similarity patterns across attention heads and preserves highly coherent groups for joint compression. This approach also eliminates the need to manually specify the group size, making the method more flexible across different models. To achieve this, we formulate the process as a clustering problem, resulting in groups with varying numbers of heads. 

Since the group proportions $\alpha_i$ are determined automatically by the clustering process, the remaining task is to allocate the ranks $r_i$. A straightforward strategy is to estimate the importance of each group based on its low-rank properties. Motivated by the fact that singular values quantify the information captured by a low-rank representation, we measure group importance using the sum of squared singular values, commonly referred to as the energy. Accordingly, the rank $r_i$ of the $i$-th group is initialized to be proportional to the energy and rounded to the nearest integer as
\[r_i = \text{round}\left(r\dfrac{\text{energy
of } W_i}{\text{energy
of } W^{(k)}}\right).\]

To ensure our approach offers a promising trade-off between structural compression and model accuracy, we constrain the total number of parameters used to represent each $W^{(k)}$ to be no greater than that of ReCalKV. Based on the analysis above, this requirement leads to
\[\sum_{i=1}^K r_i\alpha_i \le \dfrac{4r}{h}.\]

Nevertheless, the initialized $r_i$ does not necessarily guarantee this constraint. This motivates an idea of adjusting the ranks $r_i$ using a greedy heuristic. To limit energy loss, we prioritize reducing the rank of groups where each rank reduction results in the least energy loss. We therefore adopt a greedy strategy that iteratively decreases the rank of the least important group until the constraint $\sum_{i=1}^K r_i\alpha_i \le 4r/h$ is satisfied. Specifically, at each iteration, we evaluate the energy loss caused by reducing $r_i$ by one, denoted as $\Delta E_i$. Since reducing $r_i$ by one decreases $\sum_{i=1}^K r_i\alpha_i$ by $\alpha_i$, we select the group with the smallest normalized cost $\Delta E_i / \alpha_i$ at each iteration. 
Note that after this greedy adjustment, the sum of $r_i$ is no longer equal to $r$, implying that the final parameter count for each $W^{(k)}$ must be written as
\[\begin{aligned}
\sum_{i=1}^K \left[ (m \cdot r_i) + (r_i \cdot h_i d_h) \right] = m\sum_{i=1}^K r_i + n\left(\sum_{i=1}^K r_i\alpha_i\right).
\end{aligned}\]

A remaining question is how to determine the number of clusters used in the clustering algorithm. Rather than relying on conventional clustering criteria, such as the Silhouette score or the Elbow method, we propose a heuristic objective considering reconstruction error and rank utilization. In particular, for each candidate clustering configuration, after obtaining all $r_i$, the reconstruction error is simply measured as sum of the squared Frobenius norms of the difference matrix between $W_i$ and $L_i R_i$. As noted above, since all $r_i$ are modified to satisfy the parameter budget, the number of ranks used may be smaller than the available rank budget $r$. Thus, we introduce a rank utilization penalty $r - \sum_{i=1}^K r_i$. Consequently, the optimal clustering configuration is selected by minimizing the objective function
\[\sum_{i=1}^K \|W_i - L_iR_i\|_F^2 + \lambda\left(r - \sum_{i=1}^K r_i\right),\]
where $\lambda$ is a balancing coefficient to prevent either term from dominating the other. In our experiments, both terms are observed to be comparable, leading us to set $\lambda=1$ in the final configuration. 

Consequently, \autoref{algo:key_compression} summarizes the complete key compression pipeline for a Key projection matrix $W^{(k)} \in \mathbb{R}^{m \times n}$. Moreover, since all the computations are performed only once during the offline compression stage, they introduce no additional inference overhead compared with ReCalKV.

\subsection{Value compression}

Guided by ReCalKV \cite{yan2025recalkv}, we compress the Value cache by applying SVD to the original projection matrix $W^{(v)} \in \mathbb{R}^{m \times n}$ rather than processing in groups. The decomposed approximation is then represented as $W^{(v)} \approx L_vR_v$, where $L_v \in \mathbb{R}^{m \times r}$, $R_v \in \mathbb{R}^{r \times n}$ and $r$ is the rank budget. To quantify the reconstruction quality, the approximation error on a calibration set $X$ is evaluated as
\[\varepsilon = \|L_vR_vX - W^{(v)}X\|_F^2.\]

The previous allocation step implies that the Value projection matrix contains substantially higher Fisher Information than the Key projection matrix. Therefore, minimizing its reconstruction error is crucial for preserving model performance. Although SVD provides an optimal low-rank approximation in terms of the Frobenius norm, it is not necessarily optimal with respect to the activations induced by the calibration data. This indicates the need to further refine the decomposed matrices using the calibration set $X$. Specifically, the left factor $L_v$ is adjusted first. By setting the gradient of $\varepsilon$ with respect to $L_v$ to zero, we obtain
\[L_v = W^{(v)}XX^{\top}R_v^{\top} \left(R_vXX^{\top}R_v^{\top}\right)^{-1}.\]

After that, we update the right factor $R_v$ by minimizing the same objective. Setting the gradient of $\varepsilon$ with respect to $R_v$ to zero yields
\[R_v = \left(L_v^{\top}L_v\right)^{-1} L_v^{\top}W^{(v)}.\] 

These two closed-form solutions for $L_v$ and $R_v$ lead to a lower reconstruction error, result in a more faithful low-rank approximation of the original Value projection matrix $W^{(v)}$.

\section{Experiments} 
\subsection{Experimental Settings}

\textbf{Implementation Details.} We adopt the official Palu \cite{chang2024palu} repository as our implementation backbone. Given that ReCalKV \cite{yan2025recalkv} is also built on top of Palu, we first re-implement ReCalKV from this codebase to establish a consistent baseline, before extending it to incorporate our proposed method. 
For a fair comparison, we use a group size of 4 when reproducing ReCalKV, following its original experimental setup. All experiments are conducted on a single NVIDIA T4 GPU.

\textbf{Models.} We evaluate our method on three instruction-tuned language models with different attention architectures to assess its generality across settings. Specifically, we consider Llama-3.2-1B-Instruct, a Grouped-Query Attention (GQA) model with 8 attention heads per group, Qwen1.5-1.8B-Chat and SmolLM2-1.7B-Instruct, which use Multi-Head Attention (MHA) with 16 and 32 attention heads, respectively. This allows us to evaluate whether our proposed strategy is effective under both modern efficient attention GQA and standard full-attention MHA architectures.

\textbf{Datasets.} We utilize WikiText-2 as our calibration dataset in both the compression ratio allocation and value compression steps. In the evaluation phase, we assess zero-shot accuracy across six QA benchmarks (OpenBookQA, HellaSwag, PIQA, ARC-e, ARC-r, and Winogrande) to evaluate the general knowledge and reasoning capabilities after KV cache compression. Furthermore, we adopt eight datasets from the LongBench benchmark (TriviaQA, Qasper, TREC, SAMSum, LCC, RepoBench-P, QMSum, and MultiNews) to examine the effectiveness of our proposed strategies in handling long-context tasks.

\subsection{Results}

\textbf{Scope and Limitations.} We acknowledge that our experimental settings are limited by the available GPU resources, which is the primary reason we evaluate only models with a relatively small number of parameters. Another limitation is that we do not report the long-context evaluation results for SmolLM2-1.7B-Instruct because all three methods - Palu, which serves as the foundation for ReCalKV and our proposed approach, ReCalKV, and our method - achieve nearly zero performance on the LongBench benchmark, making the comparison under this setting uninformative. 


\begin{table}[h]
\centering
\caption{Comparison of K Strategy and K Parameters between ReCalKV and DynaCalKV.}
\label{tab:k-params}
\renewcommand{\arraystretch}{1.2}
\begin{tabular}{l r}
\toprule
\textbf{K strategy} & \textbf{K params} \\  
\midrule
\rowcolor{gray!15} \multicolumn{2}{c}{\textbf{Llama-3.2-1B-Instruct}} \\
ReCalKV     & 3,612,672 \\
DynaCalKV & 2,953,920 \\
\textit{Difference} & \textit{-658,752 (18.23\%)} \\
\midrule
\rowcolor{gray!15} \multicolumn{2}{c}{\textbf{Qwen1.5-1.8B-Chat}} \\
ReCalKV     & 29,982,720 \\
DynaCalKV & 25,184,384 \\
\textit{Difference} & \textit{-4,798,336 (16.00\%)} \\
\midrule
\rowcolor{gray!15} \multicolumn{2}{c}{\textbf{SmolLM2-1.7B-Instruct}} \\
ReCalKV     & 33,988,608 \\
DynaCalKV & 11,818,176 \\
\textit{Difference} & \textit{-22,170,432 (65.23\%)} \\
\bottomrule
\end{tabular}
\end{table}

\textbf{Parameter reduction.} Since we remain the same strategy for the Value cache, we focus our analysis on the Key cache. Our theoretical analysis above shows that the proposed method always reduces the number of parameters required for the Key cache. In this section, we quantify the exact amount of this reduction. \autoref{tab:k-params} shows that the parameter reduction is relatively modest for Qwen1.5-1.8B-Chat ($16.00\%$), whereas larger reductions are achieved for Llama-3.2-1B-Instruct ($18.23\%$) and SmolLM2-1.7B-Instruct ($65.23\%$). These results suggest that the magnitude of the reduction depends on the model architecture and can be attributed to the difficulty of grouping attention heads with a larger head dimension. Qwen1.5-1.8B-Chat uses a head dimension of $d_h = 128$, while both Llama-3.2-1B-Instruct and SmolLM2-1.7B-Instruct use a head dimension of $d_h = 64$. 

\begin{figure*}[t]
    \centering
    \includegraphics[width=0.85\linewidth]{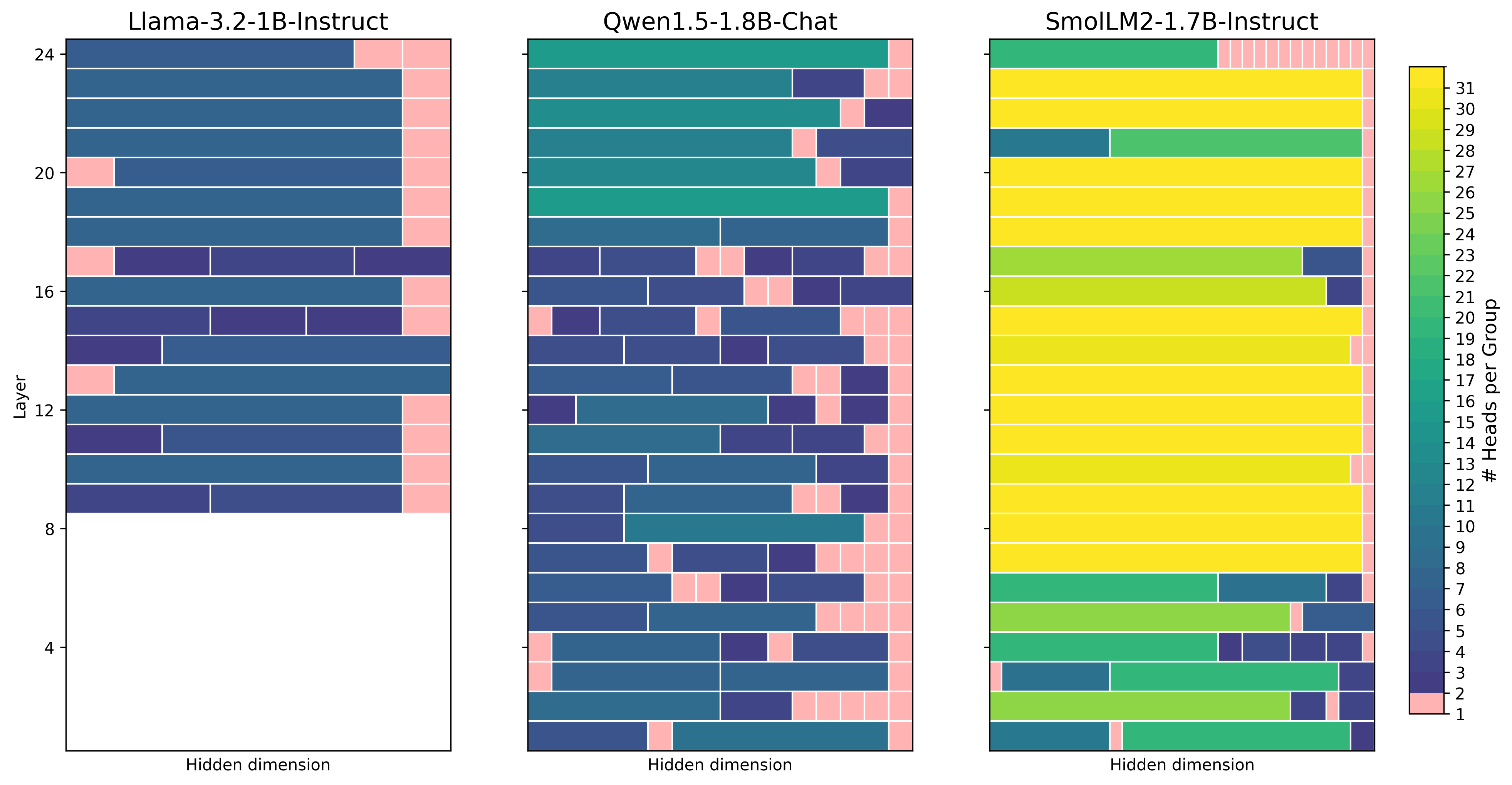}
    \caption{Visualization of attention head grouping structures across different models. Each block represents a group, where the block width corresponds to the group dimension and the color indicates the number of heads within the group. While Qwen1.5-1.8B-Chat shows a high number of 59 singleton groups, Llama-3.2-1B-Instruct and SmolLM2-1.7B-Instruct record only 17 and 38, respectively.}
    \label{fig:structure}
\end{figure*}

The higher-dimensional head representations tend to be less similar, making the Agglomerative Clustering algorithm less likely to merge multiple heads into the same group. As a result, more singleton groups are retained in Qwen1.5-1.8B-Chat, which is also illustrated in \autoref{fig:structure}. Since singleton groups preserve more distinctive information, they are less affected by the $r_i$ adjustment step, resulting in only a small parameter reduction. In contrast, larger groups contain more redundant heads and are allowed to sacrifice more rank components during the adjustment step, leading to a larger reduction.


Regarding the lower-dimensional head representations, although the reduction on Llama-3.2-1B-Instruct is much smaller than that on SmolLM2-1.7B-Instruct, this is mainly because the former employs only 16 Key projection layers with a total dimension of 512, while the latter utilizes 24 Key projection layers with a total dimension of 2048. Since each Key projection layer contributes to the parameter reduction, having fewer such layers naturally leads to a smaller total reduction. Furthermore, this difference is also magnified by their head configurations, 8 heads with a 512 total dimension for Llama versus 32 heads with a 2048 total dimension for SmolLM2. Following the same mechanism detailed above, the fewer heads in Llama are more likely to form smaller clusters, whereas the 32 heads in SmolLM2 tend to form larger groups, allowing them to sacrifice more rank components during the adjustment step.

\begin{table*}[t]
\centering
\caption{Zero-Shot Accuracy (\%) Performance Across Six Standard Benchmarks Comparing ReCalKV and DynaCalKV.}
\label{tab:results-1}
\begin{tabular}{l c c c c c c c c}
\toprule
\textbf{K strategy} & \textbf{OpenBookQA} & \textbf{HellaSwag} & \textbf{PIQA} & \textbf{ARC-e} & \textbf{ARC-r} & \textbf{Winogrande} & \textbf{\textit{Average}}\\  
\midrule
\rowcolor{gray!15} \multicolumn{8}{c}{\textbf{Llama-3.2-1B-Instruct}} \\
\midrule
    ReCalKV     & 19.60 & 36.22 & 64.47 & 53.79 & 26.71 & 52.88 \\
    DynaCalKV   & 19.20 &  35.63 & 65.07 & 54.17 & 26.19 & 52.49 \\
    \textit{Difference} & -0.40 & -0.59 & \textbf{+0.60} & \textbf{+0.38} & -0.52 & -0.39 & -0.1533 \\
\midrule
\rowcolor{gray!15} \multicolumn{8}{c}{\textbf{Qwen1.5-1.8B-Chat}} \\
\midrule
    ReCalKV     & 25.40 & 44.18 & 71.16 & 63.09 & 32.76 & 58.64 \\
    DynaCalKV & 22.80 &  43.90 & 71.76 & 63.64 & 30.97 & 57.30 \\
    \textit{Difference} & -2.60 & -0.28 & \textbf{+0.60} & \textbf{+0.55} & -1.79 & -1.34 & -0.8100 \\
\midrule
\rowcolor{gray!15} \multicolumn{8}{c}{\textbf{SmolLM2-1.7B-Instruct}} \\
\midrule
    ReCalKV     & 24.20 & 29.90 & 58.05 & 35.94 & 26.79 & 53.28 \\
    DynaCalKV & 21.20 &  34.62 & 58.92 & 38.38 & 25.77 & 51.93 \\
    \textit{Difference} & -3.00 & \textbf{+4.72} & \textbf{+0.87} & \textbf{+2.44} & -1.02 & -1.35 & +0.4433 \\
\bottomrule
\end{tabular}
\end{table*}

\begin{table*}[t]
\centering
\caption{Performance Evaluation on LongBench Datasets Comparing ReCalKV and DynaCalKV.}
\label{tab:results-2}
\begin{tabular}{l c c c c c >{\centering\arraybackslash}p{2cm} c c c}
\toprule
\textbf{K strategy} & \textbf{TriviaQA} & \textbf{Qasper} & \textbf{TREC} & \textbf{SAMSum} & \textbf{LCC} & \textbf{RepoBench-P} & \textbf{QMSum} & \textbf{MultiNews} & \textbf{\textit{Average}}\\ 
\midrule
\rowcolor{gray!15} \multicolumn{10}{c}{\textbf{Llama-3.2-1B-Instruct}} \\
\midrule
    ReCalKV     & 56.40 & 15.10 & 33.00 & 24.84 & 22.24 & 27.86 & 17.33 & 19.19 \\
    DynaCalKV & 37.44 & 06.00 & 20.50 & 14.34 & 21.67 & 24.85 & 15.15 & 11.79 \\
    \textit{Difference} & -18.96 & -9.10 & -12.50 & -10.50 & -0.57 & -3.01 & -2.18 & -7.40 & -8.03 \\
\midrule
\rowcolor{gray!15} \multicolumn{10}{c}{\textbf{Qwen1.5-1.8B-Chat}} \\
\midrule
    ReCalKV   & 68.37 & 17.20 & 42.00 & 33.01 & 28.59 & 30.44 & 17.04 & 23.68 \\
    DynaCalKV & 67.52 & 15.35 & 38.00 & 33.72 & 27.15 & 29.62 & 17.26 & 23.53 \\
    \textit{Difference} & -0.85 & -1.85 & -4.00 & +0.71 & -1.44 & -0.82 & +0.22 & -0.15 & -1.02 \\
\bottomrule
\end{tabular}
\end{table*}


\textbf{Zero-shot Accuracy Evaluation.} \autoref{tab:results-1} reports the zero-shot performance across six standard benchmarks. The empirical results suggest that while our strategy drastically downscales the Key cache parameter budget, it maintains highly competitive accuracy compared to ReCalKV. Specifically, the metrics highlight two distinct behaviors based on the characteristics of the tasks. On common knowledge tasks, such as PIQA and ARC-e, DynaCalKV consistently yields performance improvements over ReCalKV across all three models, achieving gains of $+0.87\%$ on PIQA and $+2.44\%$ on ARC-e. On more complex reasoning tasks, such as OpenBookQA, HellaSwag, ARC-r, and Winogrande, DynaCalKV experiences only minor degradation, with performance drops generally remaining within $0.28\%$ to $3.00\%$. This suggests that grouping redundant Key cache does not significantly harm the retrieval of fundamental factual knowledge, but may adversely affect the reasoning procedure. Another phenomenon is observed in SmolLM2-1.7B-Instruct, where DynaCalKV outperforms ReCalKV on HellaSwag by a notable $+4.72\%$ increase, contributing to a positive average gain of $+0.4433\%$ across all six benchmarks.


\textbf{LongBench Evaluation.} For Qwen1.5-1.8B-Chat, DynaCalKV demonstrates remarkable robustness, incurring only a negligible average performance degradation of $-1.02\%$ compared to ReCalKV, while even achieving slight gains on summarization tasks such as $+0.71$ on SAMSum and $+0.22$ on QMSum. In contrast, Llama-3.2-1B-Instruct experiences a substantial drop in performance, with an average decline of $-8.03$ score. This degradation is evident across both retrieval-heavy tasks ($-18.96$ on TriviaQA and $-12.50$ on TREC) and long-document summarization ($-10.50$ on SAMSum and $-7.40$ on MultiNews). These long-context scenarios heavily expose the architectural difference between MHA and GQA. Standard MHA models retain many independent Key heads, allowing DynaCalKV to cluster redundant heads without losing functionality. Conversely, Llama-3.2-1B-Instruct already employs GQA with only 8 Key heads. Applying further head clustering on such GQA models over-compresses these few heads, causing critical loss of the fine-grained positional and contextual features required for long-context retrieval and summarization.

\textbf{Practical Implications.} Our findings suggest that DynaCalKV should be viewed as an architecture-aware compression strategy:

\begin{itemize}    
    \item \textbf{Ideal Case:} DynaCalKV excels on standard Multi-Head Attention models with high head counts, achieving massive Key cache reductions while preserving or even enhancing performance across both short- and long-context tasks.
    
    \item \textbf{Cautionary Case:} On models already using Grouped-Query Attention with few Key heads, while DynaCalKV remains effective for short-context tasks, it should be applied conservatively in long-context scenarios.
\end{itemize}

\section{Conclusion}

In this paper, we presented an improved low-rank KV cache compression framework for efficient inference in Large Language Models. Unlike previous approaches that rely on fixed attention-head grouping, our method DynaCalKV dynamically groups Key attention heads based on CKA similarity and allocates the rank budget adaptively while reducing the overall parameter budget. For the Value cache, we adopt the offline calibration strategy of ReCalKV to further improve reconstruction quality after low-rank decomposition. Experimental results on three instruction-tuned LLMs reveal that the effectiveness of DynaCalKV depends heavily on the underlying architecture: it is particularly well-suited for standard MHA architectures, whereas it should be applied more conservatively on models already utilizing GQA during long-context tasks.


\bibliographystyle{IEEEtran}
\bibliography{references}

\end{document}